\documentclass[runningheads]{llncs}
\usepackage[T1]{fontenc}
\usepackage{graphicx}
\usepackage{booktabs}
\usepackage{multirow}
\usepackage{url}

\usepackage[unicode,pdfencoding=auto,psdextra,hidelinks]{hyperref}
\usepackage{hyperxmp}

\hypersetup{%
  pdftitle={Goal-Conditioned Decision Transformer for Multi-Goal Offline Reinforcement Learning},%
  pdfauthor={Paweł Gajewski; Dominik Żurek; Marcin Pietroń; Kamil Faber},%
  pdfsubject={Goal-conditioned offline reinforcement learning with transformer policies for multi-goal robotic manipulation},%
  pdfkeywords={offline reinforcement learning, decision transformer, multi-goal reinforcement learning, goal-conditioned policy, robotics, robotic manipulation, Franka Emika Panda, Panda-Gym},%
}

\begin{document}
\title{Goal-Conditioned Decision Transformer for Multi-Goal Offline Reinforcement Learning}
\titlerunning{Goal-Conditioned Decision Transformer}
%
\author{Paweł Gajewski\orcidID{0000-0003-0931-2476} \and
Dominik Żurek\orcidID{0000-0001-5329-1452} \and
Marcin Pietroń\orcidID{0000-0001-9357-9231} \and
Kamil Faber\orcidID{0000-0003-4221-0017}}
\authorrunning{P. Gajewski et al.}
%
\institute{Faculty of Computer Science, AGH University of Krakow, Krakow, Poland
\email{pgajewski@agh.edu.pl}}
\maketitle              
\begin{abstract}
Reinforcement learning (RL) in robotics faces significant hurdles regarding sample efficiency and generalization across varying goals. While Offline RL mitigates the need for costly online interactions, its integration with goal-conditioned policies and transformer-based architectures remains underexplored. We introduce a Goal-Conditioned Decision Transformer adapted for offline multi-goal robotics. By explicitly incorporating goal states into the sequence modeling framework, our approach efficiently solves varying tasks using only pre-collected data. We validate this method on a newly released offline dataset for the Franka Emika Panda platform. Experimental results demonstrate that our approach outperforms state-of-the-art online baselines in complex tasks and maintains robustness in sparse-reward settings, even with limited expert demonstrations. 
\end{abstract}
\section{Introduction}
\label{sec:introduction}

Reinforcement learning (RL) has driven significant progress in robotics \cite{dalal2021raps}. However, standard online algorithms require continuous environment interaction, making them impractical for real-world hardware due to safety concerns, wear and tear, and time constraints. While simulations offer a partial solution, they often suffer from the "sim-to-real" gap.

To address these limitations, \textit{Offline RL} enables agents to learn optimal policies entirely from pre-collected datasets \cite{offline}. This paradigm allows for the reuse of historical data and eliminates the need for risky online exploration. However, training on a single fixed task is insufficient for general-purpose robotics. \textit{Multi-goal RL} \cite{plappert} addresses this by requiring agents to reach varying goals within the same environment structure, necessitating policies that are explicitly conditioned on the desired outcome.

Concurrently, the \textit{Decision Transformer} (DT) \cite{mordatch} has reframed RL as a conditional sequence modeling problem, leveraging the Transformer architecture \cite{vaswani} to generate actions based on desired returns. While DTs have shown promise in standard tasks, their application to multi-goal offline scenarios remains underexplored.

In this work, we bridge these domains by introducing a Goal-Conditioned DT. We modify the standard DT architecture to explicitly process goal information, allowing it to solve multi-goal robotic tasks using only offline data. We validate this approach on a custom dataset generated using the Franka Emika Panda environment \cite{gallouedec2021pandagym}.

The contributions of this paper are threefold: (1) We extend the DT to handle multi-goal settings by incorporating explicit goal-conditioning into the input sequence; (2) We release a specialized offline RL dataset for multi-goal robotic manipulation on the Franka Emika Panda platform; and (3) We demonstrate that our approach outperforms state-of-the-art online methods (TQC+HER) and Behavioral Cloning baselines in offline settings, even under sparse reward signals.

\section{Related Work}
\label{sec:related-work}

Reinforcement learning (RL) has evolved from tabular methods to Deep RL (DRL), enabling high-dimensional continuous control \cite{sutton,mnih}. In robotic control, off-policy actor-critic methods have become the standard due to their sample efficiency. Soft Actor-Critic (SAC) \cite{sac} utilizes maximum entropy to improve stability, while Truncated Quantile Critics (TQC) \cite{tqc} further alleviates overestimation bias through distributional critics and truncation. TQC currently serves as a state-of-the-art baseline for continuous control.

\subsection{Multi-Goal RL and HER}
Standard RL typically trains for a fixed goal. To improve adaptability, Multi-Goal RL conditions the policy on a desired goal input, requiring the agent to generalize across varying targets \cite{plappert}. These environments often utilize sparse, binary rewards, making exploration difficult. Hindsight Experience Replay (HER) \cite{her} overcomes this by relabeling failed trajectories as successful ones for the specific goals that were actually achieved. The combination of TQC and HER represents a robust baseline for multi-goal robotic tasks.

\subsection{Transformers in RL}
While Transformers \cite{vaswani} dominate language and vision, their application to RL is a recent development. The DT \cite{mordatch} reframes RL as conditional sequence modeling rather than value function approximation. By conditioning an autoregressive model on desired returns, states, and actions, DT matches or exceeds offline RL baselines on standard benchmarks. Recent extensions, such as the Multi-Objective DT \cite{ghanem}, have applied this to multi-objective optimization, but the specific intersection of goal-conditioned, sparse-reward robotic control in an offline setting remains an active area of research that we address in this work.

\section{Methodology}
\label{sec:methodology}

\begin{figure}[tb]
  \centering
  \includegraphics[width=0.32\linewidth]{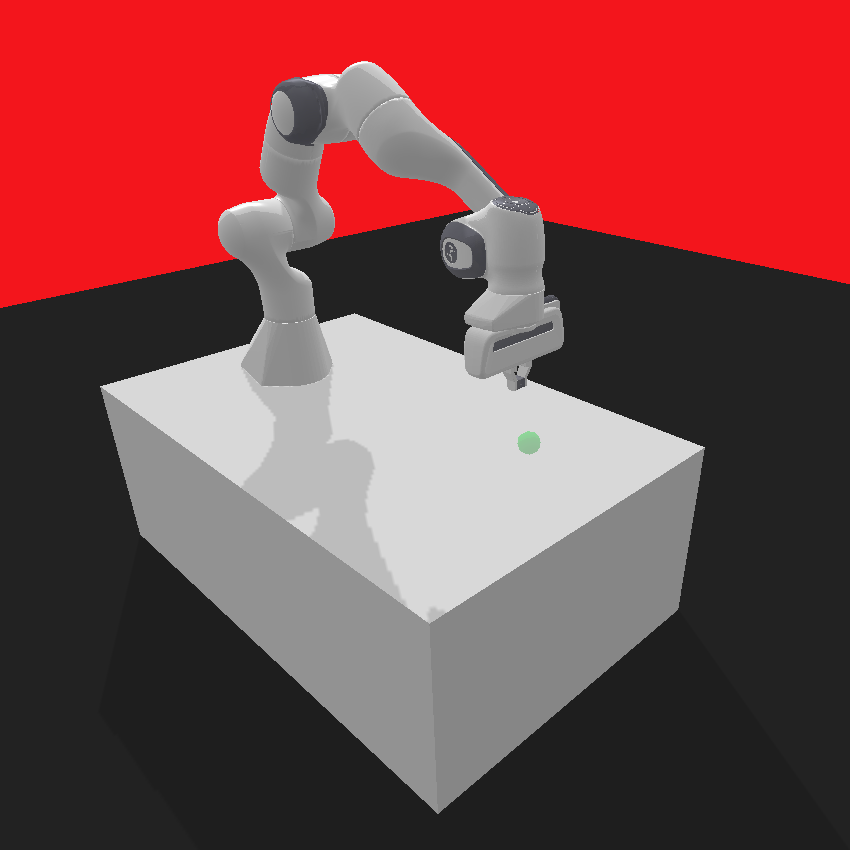}
  \includegraphics[width=0.32\linewidth]{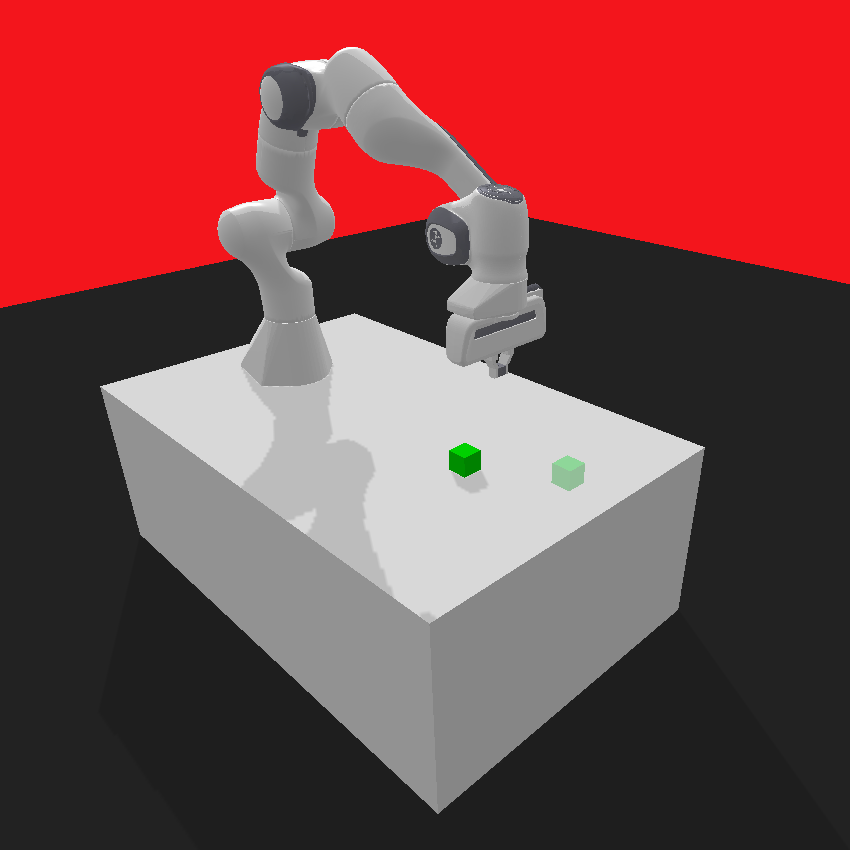}
  \includegraphics[width=0.32\linewidth]{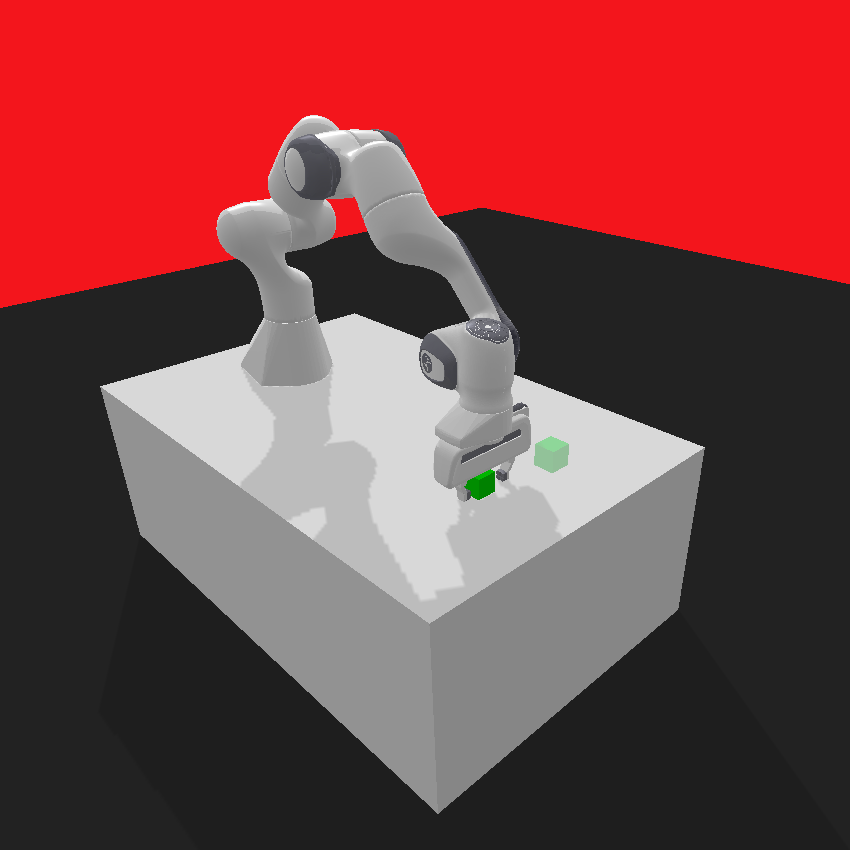}
  \caption{Multi-goal robotic environments: Reach, Push, PickAndPlace.}
  \label{fig:envs}
\end{figure}

\subsection{Decision Transformer}
We employ the DT architecture, which reframes offline reinforcement learning as a conditional sequence modeling problem. Unlike methods that fit value functions or compute policy gradients, the DT outputs optimal actions using a masked Transformer \cite{vaswani}. This architecture allows the model to effectively model trajectories by conditioning predictions on desired future outcomes.

The input sequence consists of returns-to-go $\hat{R}_t$, states $s_t$, and actions $a_t$:
\begin{equation}
\tau = \{ \hat{R}_1, s_1, a_1, \dots, \hat{R}_T, s_T, a_T \}
\end{equation}
The return-to-go at timestep $t$ is defined as $\hat{R}_t = \sum_{t'=t}^{T} r_{t'}$, representing the cumulative future reward. This conditioning enables the generation of actions required to achieve a specific target return.

\subsection{Dataset}
\label{section:dataset}
The DT is trained entirely offline. We generated two primary dataset types using the \textit{Panda-Gym} suite: \textit{expert} and \textit{random}. Expert datasets were created using converged TQC agents evaluated for 1 million timesteps. Random datasets were generated by agents sampling actions uniformly.

To evaluate data efficiency, we also created mixtures of expert and random data with varying ratios and subsets. We publicly release the dataset to support reproducibility \footnote{https://huggingface.co/datasets/lubiluk/panda-gym-offline}.

\subsection{Goal-conditioned Decision Transformer}
\label{section:methodology-dt-enhanced}

\begin{figure}[tb]
    \centering
    \includegraphics[width=0.9\linewidth]{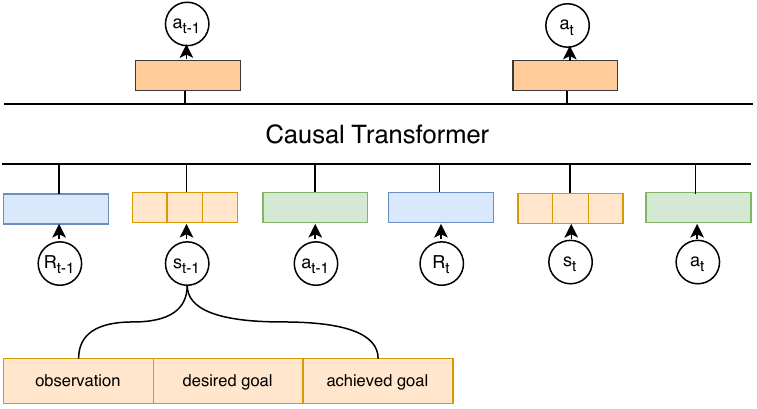}
    \caption{Goal-conditioned Decision Transformer for multi-goal RL environments}
    \label{fig:dt}
\end{figure}

We utilize multi-goal robotic environments \cite{brockman2016openai} where the observation space is a dictionary comprising the current observation $o_t$, the desired goal $g_d$, and the achieved goal $g_a$. Consequently, the state at timestep $t$ is a tuple $s_{t} = (o_t, g_d, g_a)$.

In contrast to standard DT conditioning on scalar returns, goal-conditioning provides structured, spatially meaningful supervision, which is particularly important in sparse-reward multi-goal settings where identical returns may correspond to qualitatively different objectives.

To adapt the DT to this structure, we flatten and concatenate these vectors into a single input vector:
\begin{equation}
    s_{t} = [o_t^{(0)}, \dots, o_t^{(n)}, g_d^{(0)}, \dots, g_d^{(m)}, g_a^{(0)}, \dots, g_a^{(m)}]
\end{equation}

Including the achieved goal $g_a$ allows the transformer to infer progress toward the desired goal implicitly, without requiring an explicit distance metric or reward shaping.

This modification incorporates goal-related information directly into the sequence, enabling the policy to generate actions that drive the agent toward $g_d$, as illustrated in Fig.~\ref{fig:dt}.

Unlike standard goal-conditioned policies that treat goals as part of the state input to a reactive policy, our approach conditions an autoregressive sequence model on goal information across the entire trajectory, enabling temporally extended credit assignment with respect to goal achievement.

\subsection{Tasks and Rewards} We evaluate the agent on a Franka Emika Panda robotic arm \cite{gallouedec2021pandagym} in three tasks: \textit{Reach} (position end-effector), \textit{Push} (move a cube), and \textit{PickAndPlace} (lift and move an object) illustrated by environment screenshots in Fig.~\ref{fig:envs}. We test performance under two reward structures: \textit{sparse}, where the agent receives 0 for success and -1 otherwise; and \textit{dense}, calculated as the negative Euclidean distance between the achieved goal $g_a$ and desired goal $g_d$.

A rollout is considered successful if the environment-defined success condition is met at any timestep within the evaluation horizon.

\section{Experimental setup}
\label{sec:experimental-setup}

We compare the DT against two baselines: Behavioral Cloning (BC) and Truncated Quantile Critics with Hindsight Experience Replay (TQC+HER). Our evaluation addresses five research questions: \textbf{(RQ1)} Can DT match state-of-the-art online algorithms (TQC+HER)?; \textbf{(RQ2)} How does DT compare to BC?; \textbf{(RQ3)} How does DT handle sparse vs. dense rewards?; \textbf{(RQ4)} What is the minimum dataset size required?; and \textbf{(RQ5)} How does the expert-to-random data ratio impact performance?

Experiments are conducted using 3 random seeds, with agents evaluated over 10,000 timesteps. All results are averaged over three random seeds; while limited, we observed low variance across runs. During evaluation, we condition the DT on a target return of 0 (perfect optimality) for sparse settings and 0 (zero distance) for dense settings, prompting the model to generate the most optimal trajectory. We report average return and success rate.
To address \textbf{RQ1--RQ3}, we train DT and BC on the full 1-million transition expert dataset and compare against TQC+HER trained for an equivalent number of online steps.
For \textbf{RQ4}, we evaluate data efficiency by training DT on subsets ranging from 100k to 1M expert transitions.
For \textbf{RQ5}, we analyze resilience to noise by fixing dataset size at 1M but varying the expert data ratio (0\%, 25\%, 50\%, 75\%, 100\%) mixed with random trajectories.

Hyperparameters for TQC+HER are taken from Stable-Baselines-Zoo \cite{rl-zoo3}, while DT utilizes the original paper's settings. The code is publicly available\footnote{https://github.com/lubiluk/cldt}.

While TQC+HER is trained online and thus has access to interaction during learning, we include it as a strong upper-bound baseline commonly used in robotic manipulation benchmarks.

\section{Results and Discussion}
\label{sec:results-discussion}

We present the experimental evaluation on full (1M transitions) and subset datasets, addressing the research questions formulated previously.

\begin{table*}[tb]
\caption{Performance comparison: Return / Success Rate (\%)}
\label{tab:main_results}
\centering
\scriptsize 
\setlength{\tabcolsep}{3pt} 
\begin{tabular}{lcccc}
\toprule
\textbf{Method} & \textbf{Metric} & \textbf{Reach} & \textbf{Push} & \textbf{PickAndPlace} \\
\midrule
\multicolumn{5}{c}{\textit{Dense Reward}} \\
\midrule
DT & Ret / Succ & -0.21 / 100.0 & \textbf{-0.95} / \textbf{99.5} & \textbf{-1.30} / \textbf{98.9} \\
BC & Ret / Succ & -0.21 / 100.0 & -1.20 / 95.9 & -1.35 / 97.9 \\
TQC+HER & Ret / Succ & -0.21 / 100.0 & -1.04 / 98.7 & -1.35 / 98.7 \\
\midrule
\multicolumn{5}{c}{\textit{Sparse Reward}} \\
\midrule
DT & Ret / Succ & \textbf{-1.72} / \textbf{100.0} & -8.26 / 95.0 & \textbf{-7.63} / \textbf{97.8} \\
BC & Ret / Succ & -1.76 / 100.0 & -8.18 / 94.6 & -9.01 / 94.9 \\
TQC+HER & Ret / Succ & -1.82 / 100.0 & \textbf{-4.54} / \textbf{99.5} & -16.96 / 77.0 \\
\bottomrule
\end{tabular}
\end{table*}

\begin{figure}[tb]
        \centering
        \includegraphics[width=\linewidth,trim={0 15 0 0},clip]{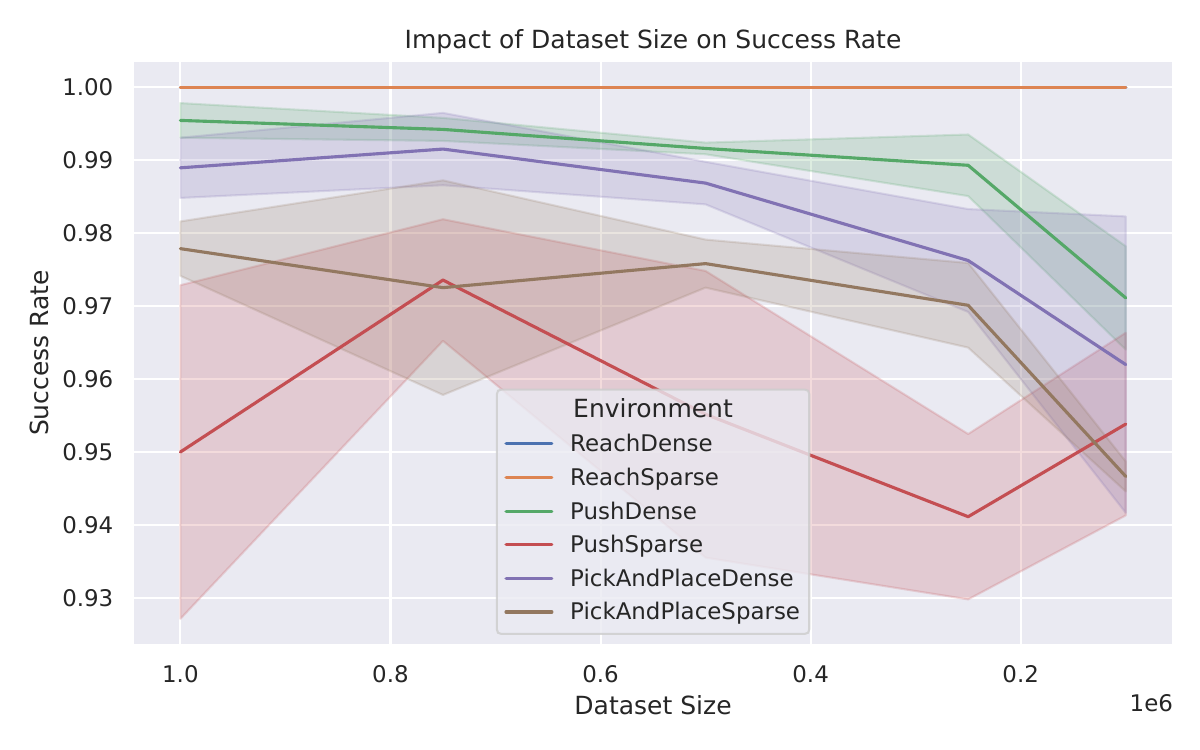}
        \caption{Impact of dataset size on success rate.}
        \label{fig:success_rate_ds_size}
\end{figure}

\begin{figure}[tb]
        \centering
        \includegraphics[width=\linewidth,trim={0 15 0 0},clip]{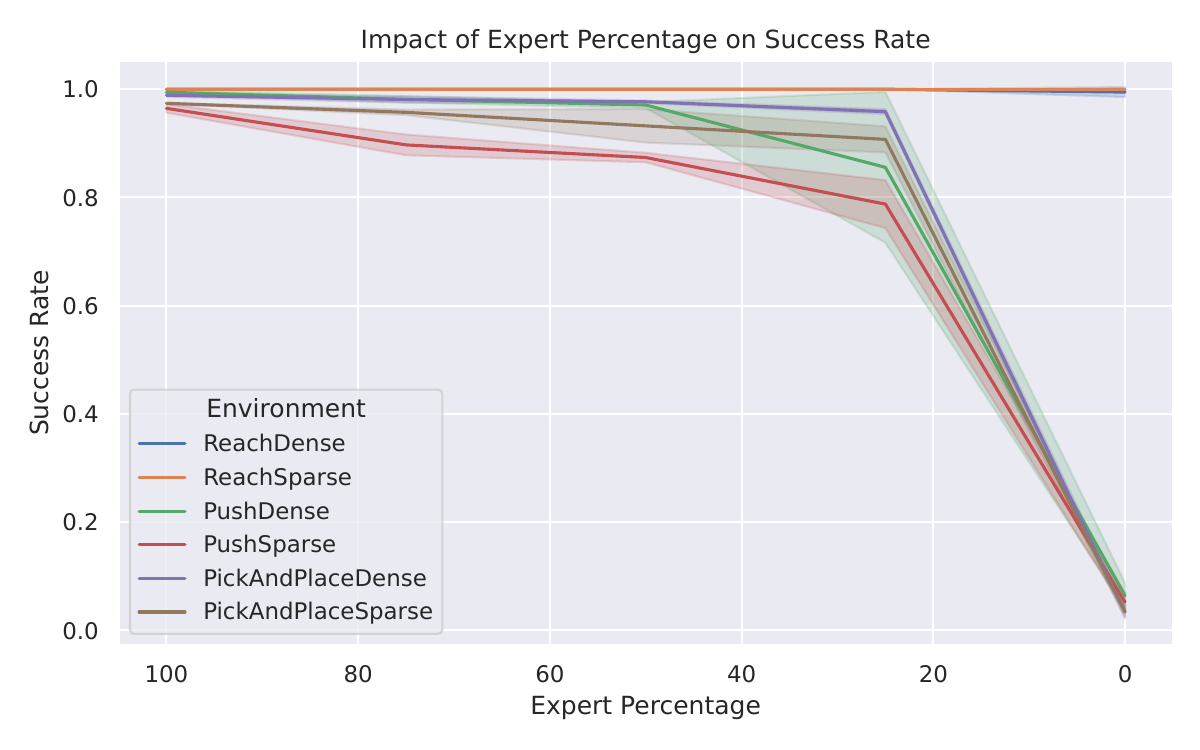}
        \caption{Impact of expert data percentage.}
        \label{fig:success_rate_expert_percentage}
\end{figure}

\subsection{Comparative Performance (RQ1, RQ2, RQ3)}
Table \ref{tab:main_results} details the performance of DT against BC and TQC+HER.
In the simplest task, \textit{Reach}, all methods achieve near-perfect success. However, in the more complex \textit{Push} and \textit{PickAndPlace} environments with Dense Rewards, DT consistently outperforms both TQC+HER and BC in average return and success rate (RQ1, RQ2). Notably, DT achieves these results with significantly lower training time (80 mins) compared to the online TQC+HER baseline (240 mins).

Under Sparse Rewards, DT demonstrates superior robustness and efficiency (RQ3). In \textit{PickAndPlace}, DT achieves a significantly higher return ($-7.63$ vs $-16.96$), indicating it solves the task in fewer steps than the baseline, while maintaining a high success rate ($97.8\%$). In contrast, TQC+HER suffers from instability in the sparse \textit{PickAndPlace} task, where its success rate drops significantly to $77.0\%$.

\subsection{Data Efficiency (RQ4)}
We analyzed the impact of dataset size on performance (Fig. \ref{fig:success_rate_ds_size}). For \textit{Reach}, performance is invariant to size. In dense reward settings for \textit{Push} and \textit{PickAndPlace}, performance declines only when data falls below 250k samples; yet, even at 100k samples, DT maintains $>96\%$ success. Sparse rewards introduce higher variance and a sharper performance drop below 250k samples.

\subsection{Expert Data Ratio (RQ5)}
Fig. \ref{fig:success_rate_expert_percentage} illustrates DT's sensitivity to trajectory quality. Expert data, i.e. trajectories from a policy that achieves near-optimal performance, was mixed with ones from a random policy. While performance degrades as the ratio of expert data decreases, DT remains surprisingly robust, maintaining $>80\%$ success in complex tasks even when expert data constitutes only 25\% of the dataset. Below this threshold, performance drops sharply.

\section{Conclusions \& Future Work}
\label{sec:conclusions-future-work}


 We presented a goal-conditioned Decision Transformer adapted for offline multi-goal robotics by explicitly incorporating goal states into the input sequence. We also release a new offline dataset for reproducibility. We hypothesize that this robustness stems from the Transformer’s ability to perform long-horizon credit assignment via self-attention, which is particularly effective in sparse-reward settings. Our results demonstrate that this approach can outperform the expert policy used to generate the training data, particularly in complex tasks like PickAndPlace. DT also exhibits robustness to suboptimal demonstrations, maintaining strong performance even when up to approximately 75\% of the trajectories are random, and remains effective under sparse reward settings. Future work will extend this framework to continual learning settings and training with heterogeneous expert trajectories.

\begin{credits}
\subsubsection{\ackname} This work was supported by funds assigned by Polish Ministry of Science and Higher Education to AGH University of Krakow, and by PLGrid HPC infrastructure (ACK Cyfronet AGH, grant no. PLG/2025/018713).

\subsubsection{\discintname}
The authors have no competing interests to declare that are relevant to the content of this article.
\end{credits}
%
%
%
\bibliographystyle{unsrt}
\bibliography{bibliography}

@article{plappert,
	title = {Multi-Goal Reinforcement Learning: Challenging Robotics Environments and Request for Research},
author={
Plappert, Matthias and Andrychowicz, Marcin and Ray, Alex and McGrew, Bob and Baker, Bowen and Powell, Glenn and Schneider, Jonas and Tobin, Josh and  Chociej, Maciek and Welinder, Peter and Kumar, Vikash and Zaremba, Wojciech},
journal={arXiv},
year={2018},
}

@article{vaswani,
author={Vaswani, Ashish and Shazeer, Noam  and  Parmar, Niki and Uszkoreit, Jakob and  Jones, Llion and  Gomez, Aidan N and Kaiser, Lukasz and Polosukhin, Illia},
title={Attention is all you need. In Advances in Neural Information Processing Systems},
year={2017},
}

@article{mordatch,
title={Decision Transformer: Reinforcement
Learning via Sequence Modeling},
author={Chen, Lili and Lu, Kevin and Rajeswaran, Aravind and Lee, Kimin and Grover, Aditya and  Laskin, Michael and Abbeel, Pieter and Srinivas, Aravind and Mordatch, Igor},
year={2022},
}

@article{ghanem,
title={Multi-Objective Decision Transformers for Offline Reinforcement Learning},
author={Ghanem, Abdelghani and Ciblat, Philippe and Ghogho, Mounir},
year={2023},
}

@article{gallouedec2021pandagym,
title        = {{panda-gym: Open-Source Goal-Conditioned Environments for Robotic Learning}},
author       = {Gallou{\'e}dec, Quentin and Cazin, Nicolas and Dellandr{\'e}a, Emmanuel and Chen, Liming},
year         = 2021,
journal      = {4th Robot Learning Workshop: Self-Supervised and Lifelong Learning at NeurIPS},
}

@article{brockman2016openai,
  title={OpenAI Gym},
  author={Brockman, G},
  journal={arXiv},
  year={2016}
}

@article{tqc,
author={Kuznetsov, Arsenii and  Shvechikov, Pavel and Grishin, Alexander and Vetrov, Dmitry},
title={Controlling Overestimation Bias with Truncated Mixture of Continuous Distributional Quantile Critics}, 
journal={ICML'20: Proceedings of the 37th International Conference on Machine Learning},
pages={5556-5566},
}

@article{her,
  title={Hindsight experience replay},
  author={Andrychowicz, Marcin and Wolski, Filip and Ray, Alex and Schneider, Jonas and Fong, Rachel and Welinder, Peter and McGrew, Bob and Tobin, Josh and Pieter Abbeel, OpenAI and Zaremba, Wojciech},
  journal={Advances in neural information processing systems},
  volume={30},
  year={2017}
}

@article{sac,
title={Soft Actor-Critic: Off-Policy Maximum Entropy Deep Reinforcement Learning with a Stochastic Actor},
author={Haarnoja, Tuomas and Zhou, Aurick and Abbeel, Pieter and Levine, Sergey},
journal={ICML},
pages={1856-1865},
year={2018},
}

@article{sutton,
title={Reinforcement Learning: An Introduction},
author={Sutton, Richard and Barto, Andrew},
year={1979},
}

@article{mnih,
title={Playing Atari with Deep Reinforcement Learning},
author={Mnih, Volodymyr and Kavukcuoglu, Koray and Silver, David and Graves, Alex and  Antonoglou, Ioannis and Wierstra, Daan and  Riedmiller, Martin},
journal={NIPS},
year={2013},
}

@inproceedings{dalal2021raps,
    Author = {Dalal, Murtaza and Pathak, Deepak and
              Salakhutdinov, Ruslan},
    Title = {Accelerating Robotic Reinforcement Learning
      via Parameterized Action Primitives},
    Booktitle = {NeurIPS},
    Year = {2021}
}

@InProceedings{offline,
  title = 	 {An Optimistic Perspective on Offline Reinforcement Learning},
  author =       {Agarwal, Rishabh and Schuurmans, Dale and Norouzi, Mohammad},
  booktitle = 	 {Proceedings of the 37th International Conference on Machine Learning},
  pages = 	 {104--114},
  year = 	 {2020},
  editor = 	 {III, Hal Daumé and Singh, Aarti},
  volume = 	 {119},
  series = 	 {Proceedings of Machine Learning Research},
  month = 	 {13--18 Jul},
  publisher =    {PMLR}
}

@misc{rl-zoo3,
  author = {Raffin, Antonin},
  title = {RL Baselines3 Zoo},
  year = {2020},
  publisher = {GitHub},
  journal = {GitHub repository},
  howpublished = {\url{https://github.com/DLR-RM/rl-baselines3-zoo}},
}

\end{document}